\documentclass[10pt,twocolumn,letterpaper]{article}

\usepackage{wacv}              %

\usepackage{graphicx}
\usepackage{amsmath}
\usepackage{amssymb}
\usepackage{booktabs}
\usepackage{xcolor}
\usepackage{algorithm}
\usepackage{algpseudocode}
\usepackage[accsupp]{axessibility}

\usepackage{mypackages}
\usepackage{mydefs}

\usepackage[pagebackref,breaklinks,colorlinks]{hyperref}

\usepackage[capitalize]{cleveref}
\crefname{section}{Sec.}{Secs.}
\Crefname{section}{Section}{Sections}
\Crefname{table}{Table}{Tables}
\crefname{table}{Tab.}{Tabs.}

\def\wacvPaperID{13} %
\def\confName{WACV}
\def\confYear{2024}

\begin{document}

\title{Tracking Skiers from the Top to the Bottom}

\author{Matteo Dunnhofer \and
Luca Sordi \and
Niki Martinel \and
Christian Micheloni \and
Machine Learning and Perception Lab, University of Udine, Udine, Italy
}

\maketitle

\begin{abstract}
   Skiing is a popular winter sport discipline with a long history of competitive events. In this domain, computer vision has the potential to enhance the understanding of athletes' performance, but its application lags behind other sports due to limited studies and datasets. This paper makes a step forward in filling such gaps. A thorough investigation is performed on the task of skier tracking in a video capturing his/her complete performance. Obtaining continuous and accurate skier localization is preemptive for further higher-level performance analyses. To enable the study, the largest and most annotated dataset for computer vision in skiing, \datasetname, is introduced. Several visual object tracking algorithms, including both established methodologies and a newly introduced skier-optimized baseline algorithm, are tested using the dataset. The results provide valuable insights into the applicability of different tracking methods for vision-based skiing analysis. \datasetname, code, and results are available at \datasetlink.
\end{abstract}

\section{Introduction}

Skiing is a globally recognized and highly popular winter sport discipline \cite{SMTreport}, with a long history of competitive events dating back to the 1840s \cite{IOCski}. 
It has been a significant part of the Winter Olympic Games since their inception in 1924 \cite{IOCski}. 
Today, skiing comprehends various disciplines such as alpine skiing, ski jumping, and freestyle skiing.  \cite{FISSKI}. Their competitive form holds a prominent position within the winter sports industry, generating over 1.7 billion media views during a winter season \cite{nielsenreportal,nielsenreportjp,nielsenreportfs}.

\begin{figure}[t]
\centering
  \includegraphics[width=\columnwidth]{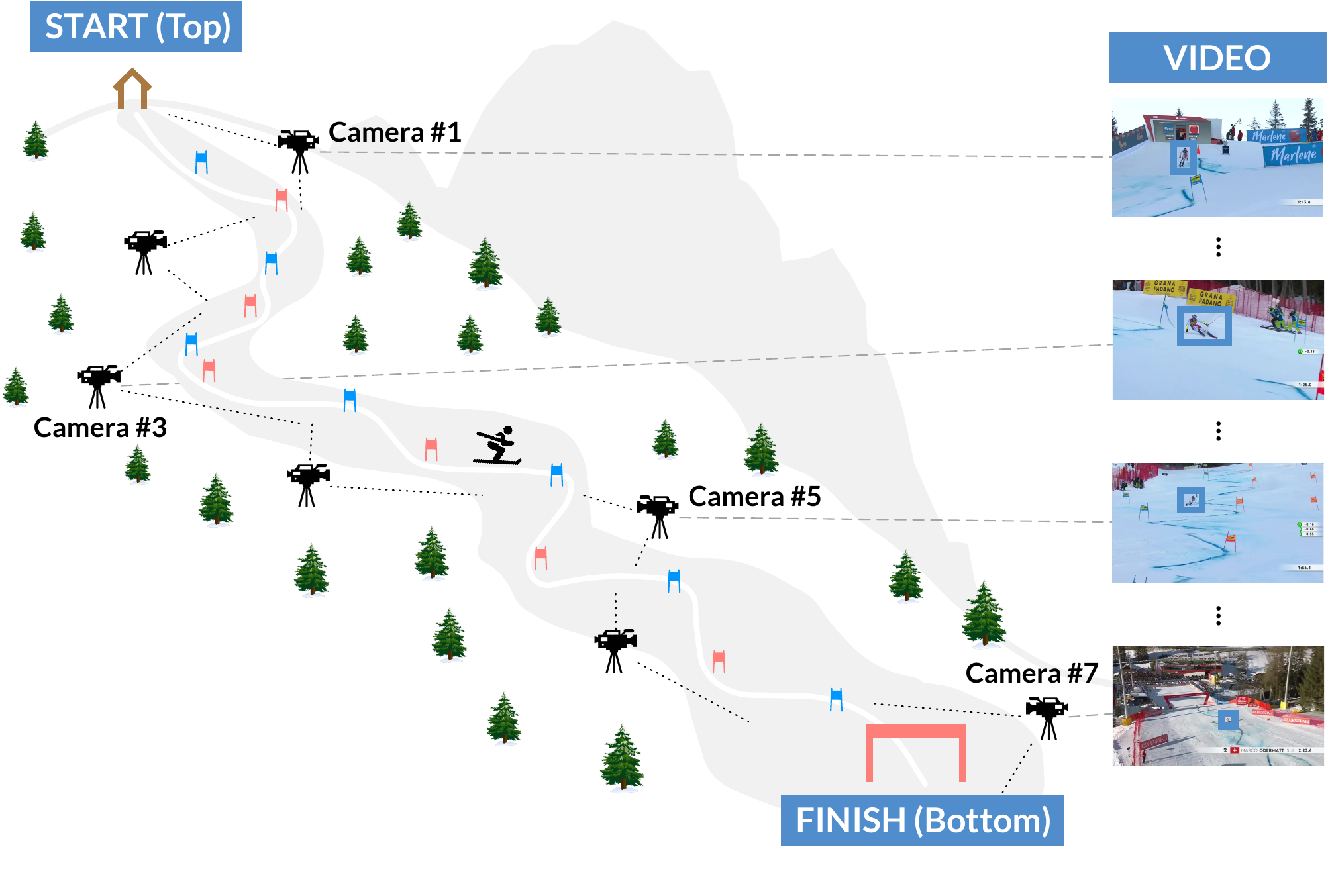}
  \caption{\textbf{Tracking a skier from the top to the bottom of the course.} This paper focuses on applying visual object tracking algorithms to localize a skier per-frame (e.g. with bounding-boxes $\textcolor[HTML]{4689cc}{\square}$) in a video capturing his/her complete performance. 
  Due to the large spatial extent of skiing courses, multiple cameras (typically pan-tilt-zoom) are placed sequentially along the slope to capture the whole performance and multi-camera tracking is  required for high-level performance analysis. 
  } 
  \label{fig:problem}
\end{figure}

Applying data-driven analytics to the skiing performance can improve the technical ability and physical well-being of athletes and augment the educational value and entertainment of broadcasting recordings of professional competitions, overall resulting in more impressive, safe, and engaging competitions.
In such applications,
computer vision offers promising opportunities for capturing and analyzing skiing performances without relying on wearable sensors. 
However, the use of computer vision in skiing has been limited compared to other sports such as soccer \cite{Vandeghen_2022_CVPR,Honda_2022_CVPR,Cioppa_2022_CVPR,Theiner_2022_WACV,Gadde_2022_WACV,Theiner_2023_WACV}, basketball \cite{bettadapura2016leveraging,bertasius2017baller,li2018rehar,Quiroga_2020_CVPR_Workshops,Chappa_2023_CVPR}, or ice-hockey \cite{pidaparthy2019keep,Koshkina_2021_CVPR,Vats_2021_CVPR,Pidaparthy_2021_CVPR,Vats_2022_CVPR}. 

Previous research in this domain has primarily focused on reconstructing skiers' poses in 2D or 3D \cite{Ludwig_2022_WACV,SkiPosePTZ} and understanding the style of ski jumps \cite{Stepec_2022_WACV,wang2019ai}.
A crucial step in all these approaches involves localizing the skier within the video frames, as this is essential for the subsequent higher-level computational analysis. 
The methods usually rely on off-the-shelf or fine-tuned object detection models \cite{FasterRCNN,YOLO,SSD} without utilizing the temporal information available in the athlete's performance evolution.
Additionally, the limited and sparsely labeled datasets used in previous studies represent a significant obstacle to the widespread development and applicability of computer vision algorithms in skiing.
Skiing videos present several unique challenges characterized by exercises performed with unique body-equipment relations, at high speed, and on a continuously changing playing field subject to extreme outdoor winter weather conditions. All of these raise important questions regarding their influence on image-based systems, and addressing them in a systematic and extensive manner could have implications even for the computer vision community as a whole, as evidenced by recent activities.\footnote{2nd Workshop on Computer Vision for Winter Sports at WACV 2023 \url{https://machinelearning.uniud.it/events/CV4WS-2023}}

This paper focuses on the study of tracking a skier in monocular broadcasting videos of professional skiing competitions. A new visual tracking dataset named \datasetname\ (``Skiers from the Top to the Bottom'') is introduced to address the lack of appropriate benchmarks. It consists of 300 video recordings across the most challenging skiing disciplines. The videos cover the skiers' complete performance, from the top to the bottom of the course, as exemplified by Figure \ref{fig:problem}. Considering the large spatial extent of courses on mountain slopes, multiple cameras are placed in sequential order along the slope to capture the complete skiing performance in such videos.
Each video is densely labeled with the bounding-boxes of a single target skier, and with attributes identifying the camera ID, the visual changes that the skier undergoes, the type of skiing discipline, the athlete ID, the location of the competition, the weather conditions, as well as the parameters of the skiing performance. \datasetname\ offers multiple training and test splits, making it suitable for developing learning-based computer vision algorithms. 
We use this benchmark to extensively evaluate different tracking algorithms, including established methodologies and a newly introduced skier-optimized baseline algorithm. Standard protocols and metrics are adapted and utilized to evaluate the specific challenges of video tracking in skiing. The impact of these tracking algorithms on higher-level skiing performance understanding tasks is also investigated. 

In sum, the contributions of this paper are:
\begin{itemize}
\item A systematic and in-depth analysis of skier tracking in videos, which has not been thoroughly studied in previous works.
\item The description and release of \datasetname\, a novel benchmark dataset curated specifically for evaluating computer vision-based systems in skiing. The dataset is designed to be diverse, representative, and densely labeled.
\item \algoname, a baseline algorithm optimized for skier tracking in multi-camera videos. 
\end{itemize}

\begin{table*}[t]

\fontsize{8}{9}\selectfont
	\centering
	\caption{\textbf{Comparison of \datasetname\ with publicly available skiing-related datasets.} This table shows a comparison between some key statistics of our \datasetname\ and of other datasets for computer vision tasks available in the skiing domain. As can be noticed, ours results in the largest, most diverse, and most annotated dataset. (n/a stands for ``not annotated''.)}
	\label{tab:skidatasets}
	\begin{tabular}{l | c  c  c  c | c }
		\toprule
		Dataset & Skimovie \cite{Skimovie} & Ski2DPose \cite{SkiPosePTZ} & SkiPosePTZ \cite{SkiPosePTZ,Rhodin2018} & YouTube Skijump \cite{Ludwig_2023_WACV} & \datasetname\\

		\midrule
        Skiing application & Detection & 2D Pose Estimation & 3D Pose Estimation & 2D Pose Estimation & Tracking \\
        Per-frame annotations & \cmark\ (12.5 FPS) & n/a & n/a & n/a & \cmark\ (30 FPS) \\
        Complete performance & \cmark  & n/a & n/a & n/a & \cmark \\
        Performance parameters & n/a  & n/a & n/a & n/a & \cmark \\
        Weather annotations & n/a & n/a & n/a & n/a & \cmark \\
		\# multi-camera videos & n/a & n/a & n/a & n/a & 300 \\
            \# single-camera videos & 4 & n/a  & 36 & n/a & 2019 \\
            
		\# annotated frames & 2718 & 1982 & 20K & 2867 & 352978  \\
            
            \# skiing disciplines & 1 (AL) & 1 (AL) & 1 (AL) & 1 (JP) & 3 (AL, JP, FS) \\
            \# sub-disciplines & 1 & 4 & 1 & 2 & 11 \\
            \# athletes & n/a & 32 & 6 & 118 & 196 \\
            \# locations & 6 & 5 & 1 & 7 & 161 \\

		\bottomrule		
\end{tabular}

\end{table*}

\section{Related Work}

\paragraph{Visual Object Tracking.}
In the recent past, there has been increasing interest in developing precise and robust single object tracking (SOT) algorithms for various domains. Early trackers utilized mean shift algorithms~\cite{Comanciu2000}, key-point~\cite{Matrioska}, part-based techniques \cite{LGT}, or SVM learning~\cite{Struck}. Correlation filters gained popularity due to their fast processing~\cite{MOSSE,KCF}. More recently, deep learning-based solutions, including regression networks~\cite{GOTURN,dunnhofer2021weakly},
online tracking-by-detection methods~\cite{MDNet}, reinforcement learning-based methods~\cite{Yun2017,dunnhofer2019visual},
deep discriminative correlation filters~\cite{ATOM,DiMP}, siamese network-based trackers~\cite{SiamFC,SiamRPNpp}, and transformers \cite{Stark,MixFormer,OSTrack,ToMP}, led to higher tracking accuracy.
Long-term trackers and methods combining multiple trackers have been also explored \cite{SPLT,GlobalTrack,CoCoLoT}.

Such a progress in SOT algorithms is attributed to well-curated evaluation datasets featuring diverse object types \cite{OTB,VOT2020,VOT2021,VOT2022} and large-scale datasets for visual object tracking in generic domains~\cite{TrackingNet,GOT10k,LaSOT}. Application-centric benchmarks exist for specific domains such as drones \cite{UAV123}, high frame-rate videos \cite{NfS}, transparent objects \cite{TOTB}, and egocentric videos \cite{dunnhofer2023visual}. These benchmarks contribute to the development of accurate and reliable tracking systems in specific application scenarios.

The aforementioned datasets \cite{OTB,VOT2022,GOT10k,TrackingNet,LaSOT} lack a sufficient representation of skiing, hindering the development of effective trackers in this domain. %
To overcome this limitation, we introduce \datasetname\ as a comprehensive and well-curated benchmark for evaluating trackers on skiing regardless of their methodology. The dataset covers the unique aspects of the skiing domain, including fast human motion, extreme weather conditions, and distractor objects. We believe that \datasetname\ can also benefit the development of generic tracking methodologies.

\paragraph{Applications of Computer Vision to Skiing.}
Recent advancements in computer vision \cite{ResNet,FasterRCNN,OpenPose} have enabled vision-based applications in skiing performance analysis. For example, 
 \cite{Zhu2022} proposed object detection and human pose estimation algorithms to recognize falls of alpine skiers, while \cite{Zwolfer2021} discussed the combination of pose estimation with kinematics models. \cite{SkiPosePTZ} introduced a methodology to reconstruct 3D poses from images captured by multiple cameras observing a single slope section.
 Ski jumping analysis involved scoring the style of jumps using 2D human pose trajectories \cite{Stepec_2022_WACV} and detecting key-points on the human body and skis in still images using improved vision transformer architectures \cite{Ludwig_2022_WACV,Ludwig_2023_WACV}. 
For freestyle skiing and snowboarding, algorithms were developed to evaluate the quality of jumps in monocular videos \cite{wang2019ai} and to synchronize videos for comparing the timing and spatial extent of aerial maneuvers \cite{Matsumura2021}.

The discussed pipelines present object detection \cite{FasterRCNN,YOLO} or off-the-shelf visual tracking  \cite{wang2019ai} for initial skier localization, followed by subsequent modules for higher-level output computation. The accuracy of skier localization greatly affects the performance of the successive modules, but this aspect has been overlooked by existing systems. Only limited evaluations on skier localization accuracy have been conducted in previous works \cite{Skimovie,qi2022alpine}. These studies focused on a small number of videos and lacked analysis of the challenging characteristics of the skiing domain. In contrast, this paper presents a systematic and comprehensive analysis of skier tracking on a large scale, involving 300 videos and 353K frames. Multi-camera videos capturing professional athletes from various skiing disciplines were used, considering real competition conditions with different courses, skiing styles, distracting skiers, and challenging weather. 
A comparison between the proposed \datasetname\ and publicly available computer vision datasets for skiing applications is presented in Table \ref{tab:skidatasets}.

\section{Problem Formulation}
\label{sec:problem}
This paper focuses on the per-frame localization of a specific skier in a video stream capturing his/her complete performance on a skiing course, from the top of the latter to its bottom. 
The video stream is a sequence $\video = \big\{ \frame_t \in \images \big\}_{t=0}^{T}$ of frames $\frame_t$, where $\images$ represents the space of RGB images and $T$ is the total number of frames. The bounding-box $\bbox_t = [x_t,y_t,w_t,h_t] \subseteq \reals^4$ defines the skier's position and size in each frame, and the objective is to develop a visual tracking algorithm -- also referred to as tracker -- to predict the bounding-box $\bbox_{t}$ with a confidence value $0 \leq \conf_t \leq 1$, for $0 < t \leq T$, in an online fashion.
The initial bounding box $\bbox_0$ can be generated by an object detection algorithm \cite{YOLO,FasterRCNN} or manually annotated by a human operator. 
Skiing competitions involve courses spanning several hundred meters if not kilometers, requiring multiple cameras to be placed sequentially along the slope to capture the skier's entire performance.
Thus, $\video$ consists of frames grabbed by several different cameras and concatenated into a single stream showing a complete performance. 
Considering that skiing is an individual sport, our problem of interest constitutes an application case of single long-term object tracking \cite{Lukezic2020,VOT2022}, specifically of the global instance variation \cite{GIT} which aims to continuously localize a target object over an extended period, even across camera shot-cuts.
Figure \ref{fig:problem} presents a visualization of such a setting for the case of alpine skiing.
We assume that manual camera control and camera switching occur, as it is done for real-time broadcasting transmission. Our paper focuses on the problem of per-frame skier localization and does not cover the broader task of tracking a skier across a network of automatically controlled cameras. We believe that our findings can  contribute to the development of such technology, since to control cameras  the athlete must be first localized in the frames of the video stream.

\section{The \datasetname\ Dataset}
The \datasetname\ dataset provides a comprehensive spatio-temporal video representation and annotation of professional skiing performance under the settings described in Section \ref{sec:problem}. \datasetname\ comes with dense annotations for tracking purposes, but it is designed to serve as a well-curated benchmark for subsequent higher-level skiing performance understanding tasks.
Particularly, we adhere to the following design principles:\footnote{Further details are given in Appendix \ref{sec:datadet} of the supp. document.}
\begin{itemize}
    \item Scale: we ensured that \datasetname\ would contain a significant number of videos and frames to facilitate the development of modern computer vision solutions based on deep learning.
    \item Diversity: we included a wide range of situations, such as different skiing disciplines, athletes, skiing styles, courses and locations, to enable the testing and generalization of methods under various application conditions.
    \item Representativeness: we designed \datasetname\ to represent real competition scenarios of professional athletes, which enables the development of algorithms capable of working in real-world situations.
\end{itemize}

\begin{table}[t]

\fontsize{8}{9}\selectfont
	\centering
	\caption{\textbf{Key statistics of \datasetname.} The following table offers overall and per-skiing discipline information about the videos and the associated data present in the proposed dataset. 
 }
	\label{tab:stats}
	\setlength\tabcolsep{.11cm}
	\begin{tabular}{l | c  c  c  | c }
		\toprule
		Skiing Discipline & AL & JP & FS & All \\
		\midrule
		\# MC videos & 100 & 100 & 100 & 300 \\
            \# SC videos & 1100 & 346 & 573 & 2019 \\
		\# frames & 215517  & 38201 & 99260 & 352978  \\
        \# cameras (min, avg, max) & (6, 11, 26) & (2, 3, 5) & (1, 6, 15)  & (1, 7, 26)\\
        avg MC video seconds & 71 & 13 & 33 & 39  \\
        avg SC video seconds & 6.5 & 3.6 & 5.7  & 5.8  \\
            \# sub-disciplines & 4 & 2 & 5 & 11 \\
            \# athletes & 56 & 54 & 86 & 196 \\
            \# athlete genders (M, W) & (34, 22) & (35, 19) & (49, 37) & (118, 78) \\
            \# athlete nationalities & 15 & 10 & 18 & 25 \\
            \# courses & 68 & 34 & 59 & 161 \\
            \# courses countries & 15 & 12 & 17 & 24 \\
		
		\bottomrule		
\end{tabular}

\end{table}

The challenge represented by \datasetname\ arises from the complex and dynamic nature of skiing and its environment, where an athlete's visual appearance and motion are influenced by highly variable factors such as: complex body movements due to high speed, course settings, aerial execution, impact absorption;  particular image characteristics due to meteorological conditions (e.g., snowing, raining, and intense shadowing) and  camera operations (e.g., camera switching, fast camera movements, long-range capturing). 
From a more general point of view, \datasetname\ can serve as a valuable resource for research in multi-camera target tracking under extremely dynamic outdoor environments.

\paragraph{Video Collection.}
The videos in \datasetname\ were carefully selected from broadcasting recordings showcasing complete skiing performances available on the Internet.
Our selection process aimed to maximize diversity in terms of athletes, locations, courses, and weather conditions, while ensuring a balanced distribution across three major skiing disciplines defined by the International Ski and Snowboard Federation (FIS) \cite{FISSKI}: alpine skiing (AL), ski jumping (JP), and freestyle skiing (FS). These disciplines were chosen based on their popularity and the challenging representations they provide in videos. Existing datasets \cite{Skimovie,SkiPosePTZ,Ludwig_2023_WACV} did not encompass all the desired characteristics, necessitating the creation of a new video collection. The videos have a framerate of 25/30 FPS and resolutions ranging from 360p to 720p. More detailed statistics are in Table \ref{tab:stats}.

\begin{figure}[t]
\centering
  \includegraphics[width=\linewidth]{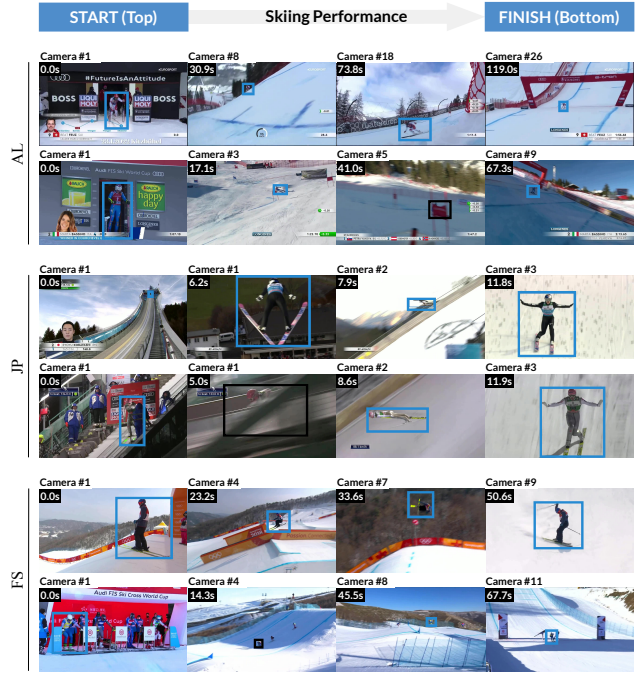}
  \caption{\textbf{Frame and bounding-box samples from \datasetname.} We showcase examples of video frames from our dataset for the different disciplines: alpine skiing (AL), ski jumping (JP), and freestyle skiing (FS). Each frame is accompanied by a manually annotated bounding-box. A blue rectangle ($\textcolor[HTML]{4689cc}{\square}$) localizes the  skier's appearance as visible, while a black rectangle ($\textcolor{black}{\square}$) as occluded. The camera that captured the frame and the elapsed time in seconds from the beginning of the performance are also reported.} 
  \label{fig:videex}
\end{figure}

\paragraph{Frame-level Annotations.}
Each of the frames belonging to the 300 multi-camera (MC) videos has been manually labeled with the bounding-box enclosing the visual appearance of the athlete and its equipment (skis, and poles if present), as shown in Figure \ref{fig:videex}. 
The sequence of boxes for each video starts with a frame capturing at least 50\% of the skier's appearance shortly before the descent begins, and it ends on a frame capturing the skier after completing their performance. 
Each box is labeled to indicate whether the skier is visible or occluded (\textit{i.e.},  when approximately more than 50\% of the skier's visual appearance is hidden).
The dataset includes instances of complete occlusions, such as when the skier passes behind snow ramps in FS. In such cases, boxes are drawn to localize the skier in likely positions based on the observed motion. On average, complete occlusions last for 15 frames. 
The motivation for employing bounding-boxes as target representations stems from the requirement, by higher-level performance understanding modules (e.g. 2D or 3D pose detectors \cite{Stepec_2022_WACV,Ludwig_2023_WACV,SkiPosePTZ}), of a rectangle-shaped localization to extract athlete-specific visual information. In this view, bounding-boxes offer a simpler yet sufficiently informative representation that provides the position and the size of the athlete's appearance in the frames.
Each frame is also labeled with the index of the camera that captured it. The camera order for each video was manually determined by assessing the order of video shot-cuts. This enumeration reflects the sequence in which the cameras were positioned along the slope.

\paragraph{Video-level and Clip-level Annotations.}
To enable in-depth analysis, we have associated labels with both the MC videos and the single-camera (SC) clips, which are sub-sequences of frames captured by the same camera. 
Each MC video is labeled with the following information: the discipline (AL, JP, FS); the specific skiing sub-discipline; the visible weather condition; %
athlete ID (including name and surname) and nationality; the date, location, and country of the competition. 
It's worth noting that each MC video is also annotated with the athlete's performance results in computable form, even though these labels are not specifically utilized in this work.
Following the established practice of visual object tracking benchmarks \cite{OTB,UAV123,VOT2022,LaSOT,dunnhofer2023visual}, each SC clip is associated with labels (CM, SC, BC, ARC, IV, POC, MB, FM, FOC, LR) that express the visual variability of the target skier, reinterpreted to suit the application domain.
All of these labels can be utilized for video clustering, enabling experimental results to be conditioned on different characteristics of the domain.

\paragraph{Training-Test Splits.}
To enable training and evaluation of machine learning-based trackers, the MC videos are divided into training and test sets, following three different split conditions each with a 60-40 ratio.
The first split follows a conventional deployment approach, where models are trained on past data and tested on newer data. This split is based on the dates associated with the videos.
The second split focuses on evaluating the models' generalization ability to unseen athletes. It involves creating separate training and test sets based on disjoint athlete IDs.
The third split assesses the models' generalization to new skiing courses. In this case, dedicated disjoint partitions are formed using the location data associated with each video.

\paragraph{Data Quality.}
The video selection and annotation process was meticulously carried out by our research team, consisting of an MSc student, a post-doc researcher, and two professors. All annotators had research experience in visual object tracking and in watching skiing competitions on TV. To ensure additional accuracy, we sought application-specific guidance from two professional alpine skiing coaches and a FIS-licensed ski jumping judge.
We utilized the CVAT tool \cite{cvat} for drawing and validating the bounding-boxes. The metadata associated with the videos, including performance parameters and weather conditions, was obtained from the publicly available FIS database \cite{FISSKI}.

\section{Trackers}

\paragraph{Trackers for Generic Objects.}
In our evaluation, we considered a range of state-of-the-art methods designed for tracking arbitrary objects, including long-term trackers specifically designed for addressing abrupt target changes and occlusions \cite{Lukezic2020}, as in our application of interest. The trackers falling in this category include SPLT \cite{SPLT}, GlobalTrack \cite{GlobalTrack}, LTMU \cite{LTMU}, KeepTrack \cite{KeepTrack}, STARK \cite{Stark}, and CoCoLoT \cite{CoCoLoT,dunnhofer2022combining}. 
In addition, we included the short-term trackers \cite{Lukezic2020} MOSSE \cite{MOSSE}, KCF \cite{KCF}, and SiamRPN++ \cite{SiamRPNpp} for their general popularity, and MixFormer \cite{MixFormer}, OSTrack \cite{OSTrack}, FEAR \cite{FEAR}, and SeqTrack \cite{SeqTrack} for their very recent demonstration of high accuracy. Although these methods were not explicitly designed for long-term tracking tasks, some of them have shown promising performance in similar conditions \cite{LaSOT} and could be even suitable for skier localization in  SC tracking tasks.

\begin{table*}[t]
\fontsize{7}{8}\selectfont
	\centering
	\caption{\textbf{Overall and per-discipline results in the multi-camera (MC) tracking setting.} The \fscore, \prec, and \recall\ scores are presented for each studied algorithm. In general, we observe that ski jumping (JP) is the discipline in which trackers perform better, followed by alpine skiing (AL). Freestyle skiing (FS)  offers the most challenging situations.
	}
	\label{tab:perdisciplinelt}
	\setlength\tabcolsep{.24cm}
	\begin{tabular}{c | c   c  c  c c c c c c c c c | c c c}
		\toprule
		\rotatebox[origin=c]{90}{Discipline} & \rotatebox[origin=c]{90}{MOSSE} & \rotatebox[origin=]{90}{KCF} & \rotatebox[origin=c]{90}{SiamRPN++} & \rotatebox[origin=c]{90}{FEAR} & \rotatebox[origin=c]{90}{GlobalTrack} & \rotatebox[origin=c]{90}{MixFormer} & 
  \rotatebox[origin=c]{90}{KeepTrack} &
  \rotatebox[origin=c]{90}{OSTrack} & \rotatebox[origin=c]{90}{SeqTrack} &  \rotatebox[origin=c]{90}{LTMU} & 
        \rotatebox[origin=c]{90}{CoCoLoT} & 
  \rotatebox[origin=c]{90}{STARK} & \rotatebox[origin=c]{90}{\yolotr} & \rotatebox[origin=c]{90}{\starkft} & \rotatebox[origin=c]{90}{\algoname} \\
		\midrule

        \multirow{3}{*}{All} & 0.093 & 0.061 & 0.248 & 0.338 & 0.493 & 0.526 & 0.527 & 0.528  & 0.534  & 0.554  & 0.562 & 0.584  & \third{0.740} & \second{0.818} & \first{0.835} \\
         & \smallresult{0.092} & \smallresult{0.061} & \smallresult{0.270} & \smallresult{0.419} & \smallresult{0.493} & \smallresult{0.518} & \smallresult{0.555} & \smallresult{0.520} & \smallresult{0.538} & \smallresult{0.565} & \smallresult{0.572} & \smallresult{0.595} & \smallresult{0.730} & \smallresult{0.832} & \smallresult{0.843}\\
         & \smallresult{0.094} & \smallresult{0.062} & \smallresult{0.235} & \smallresult{0.301} & \smallresult{0.495} & \smallresult{0.535} & \smallresult{0.508} & \smallresult{0.537} & \smallresult{0.533} & \smallresult{0.545} & \smallresult{0.555} & \smallresult{0.576} & \smallresult{0.751} & \smallresult{0.806} & \smallresult{0.829}\\

         \midrule

        \multirow{3}{*}{AL} & 0.031 & 0.024 & 0.144 & 0.270 & 0.485 & 0.463 & 0.518 & 0.462 & 0.479  & 0.524 & 0.532 & 0.552 &  0.798 & 0.853 & 0.868 \\
         & \smallresult{0.031} & \smallresult{0.024} & \smallresult{0.143} & \smallresult{0.260} & \smallresult{0.487} & \smallresult{0.458} & \smallresult{0.561} & \smallresult{0.457} & \smallresult{0.485} & \smallresult{0.541} & \smallresult{0.546} & \smallresult{0.565} & \smallresult{0.790} & \smallresult{0.874} & \smallresult{0.885} \\
         & \smallresult{0.032} & \smallresult{0.024} & \smallresult{0.145} & \smallresult{0.229} & \smallresult{0.483} & \smallresult{0.468} & \smallresult{0.484} & \smallresult{0.467} & \smallresult{0.475} & \smallresult{0.509} & \smallresult{0.521} & \smallresult{0.540} & \smallresult{0.807} & \smallresult{0.834} & \smallresult{0.852} \\

        \midrule
        
        \multirow{3}{*}{JP} & 0.155 & 0.098 & 0.281 & 0.373 & 0.504 & 0.574 & 0.536 & 0.577 & 0.590  & 0.576 & 0.584 & 0.603 & 0.818 & 0.880 & 0.896 \\
         & \smallresult{0.153} & \smallresult{0.097} & \smallresult{0.310} & \smallresult{0.451} & \smallresult{0.507} & \smallresult{0.567} & \smallresult{0.576} & \smallresult{0.571} & \smallresult{0.598} & \smallresult{0.591} & \smallresult{0.606} & \smallresult{0.630} & \smallresult{0.807} & \smallresult{0.892} & \smallresult{0.898} \\
         & \smallresult{0.157} & \smallresult{0.099} & \smallresult{0.262} & \smallresult{0.338} & \smallresult{0.502} & \smallresult{0.581} & \smallresult{0.510} & \smallresult{0.584} & \smallresult{0.584} & \smallresult{0.565} & \smallresult{0.569} & \smallresult{0.582} & \smallresult{0.830} & \smallresult{0.871} & \smallresult{0.896} \\

        \midrule
        
        \multirow{3}{*}{FS} & 0.092 & 0.065 & 0.319 & 0.372 & 0.491 & 0.541  & 0.528 & 0.545 & 0.533  & 0.562 & 0.570 & 0.596 & 0.603 & 0.721 & 0.742 \\
         & \smallresult{0.090} & \smallresult{0.067} & \smallresult{0.358} & \smallresult{0.446} & \smallresult{0.483} & \smallresult{0.529} & \smallresult{0.528} & \smallresult{0.532} & \smallresult{0.530} & \smallresult{0.564} & \smallresult{0.564} & \smallresult{0.590} & \smallresult{0.592} & \smallresult{0.730} & \smallresult{0.746} \\
         & \smallresult{0.094} & \smallresult{0.080} & \smallresult{0.298} & \smallresult{0.336} & \smallresult{0.500} & \smallresult{0.556} & \smallresult{0.530} & \smallresult{0.560} & \smallresult{0.539} & \smallresult{0.562} & \smallresult{0.577} & \smallresult{0.604} & \smallresult{0.616} & \smallresult{0.713} & \smallresult{0.738} \\

		\bottomrule		
\end{tabular}
\end{table*}

\paragraph{Skier-specific Trackers.}
We also assessed the performance of  baseline trackers specifically designed for skier tracking. The \yolotr\ tracker adopts a tracking-by-detection approach utilizing an YOLOX instance \cite{YOLOX,YOLOv5} (fine-tuned on  \datasetname's training set) and the SORT \cite{SORT} algorithm to detect and associate boxes across consecutive frames.
\starkft\ instead implements an instance of STARK (STARK-ST50) \cite{Stark} fine-tuned on \datasetname\ using the original hyper-parameter values.

\textit{\algoname.} Furthermore, we introduce a new tracker, called \algoname, which consists of two instances of \starkft. The first instance serves as a precise skier tracker during periods of target visibility, typically within an SC clip where skier and camera move smoothly. The second instance acts as a skier re-detector, activated when the confidence $\conf_t$ of the first instance is low, often occurring after camera shot-cut or target occlusion.
The first instance uses a smaller factor (3.0 instead of 5.0) to compute a search area with higher resolution that includes more information of the target's visual appearance, resulting in more accurate bounding-box predictions. 
The second instance re-locates the target whenever the confidence of the first instance drops below 0.5. It uses the standard search area factor (5.0) to find the skier in a larger frame area centered at the position predicted by the first instance. Pseudo-code and further details are given in Appendix \ref{sec:skiertrackers} of the supp. document.

\section{Evaluation}

\begin{figure}[t]
\centering
  \includegraphics[width=\columnwidth]{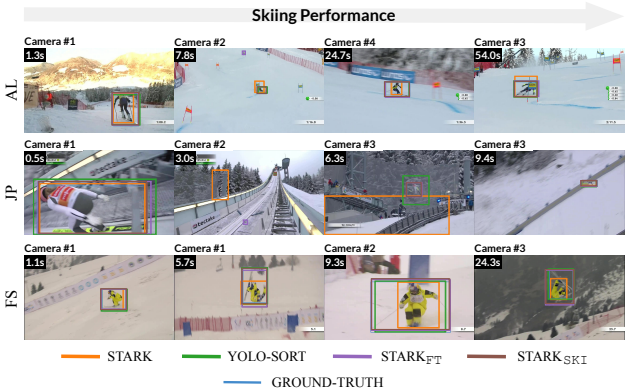}
  \caption{\textbf{Qualitative tracking performance.} This figure shows bounding-box samples predicted by the top four trackers for frames of \datasetname's test set. \starkft\ and \algoname\ exhibit high precision in localizing both the skier's body and equipment. Videos with results can be visualized at this link: \url{https://www.youtube.com/watch?v=Aos5iKrYM5o}.} 
  \label{fig:qualex}
\end{figure}

\subsection{Tracking Accuracy}

\paragraph{Evaluation Protocol.}
We utilized the one-pass evaluation (OPE) protocol \cite{OTB} to run trackers for performance assessment. This protocol closely simulates tracking implementation in real application conditions. It involves initializing the tracker with a target bounding-box in the first frame and running it on subsequent frames until the end of the video. In our case, this means initializing the tracker at the beginning of the skier's performance and running it until the end.
By default, our experiments were conducted using the ground-truth bounding-box for initialization, thus providing the best possible evaluation conditions for the trackers. By considering practical deployment scenarios where automatic athlete localization is required (such as real-time analysis during broadcasting), we also tested the trackers by initialization with the box predicted by the YOLOX detector \cite{YOLO,YOLOv5} trained on \datasetname's training set.
If not specified otherwise, in the experiments all the trackers were executed on the date-based test-set, with skier-specific trackers trained on the corresponding training set.

\paragraph{Performance Measures.}
\label{sec:trpm}
To quantify the trackers' performance, we utilized standard measures for long-term tracking evaluations \cite{Lukezic2020}: Precision (\prec), Recall (\recall), and F-Score (\fscore).
In simple terms, \prec\ measures the average number of confidently tracked ground-truth boxes, considering different Intersection-over-Union (IoU) thresholds to determine correct predictions. \recall\ measures the same number, regardless of the tracker's confidence. \fscore\ combines \prec\ and \recall\ into a single aggregated score.
We use the Generalized Robustness Score (\gsr) \cite{dunnhofer2023visual,dunnhofer2021first} to quantify the extent of consecutive frames successfully tracked before losing the target. Such an event occurs whenever the tracker does not resume correct tracking after some time.
Additionally, we evaluate the efficiency of trackers by \timediff\ which measures the time in seconds that has to be waited to obtain the skier's localization after the athlete's performance state is observed in each frame.

\begin{figure}[t]
\centering
  \includegraphics[width=\linewidth]{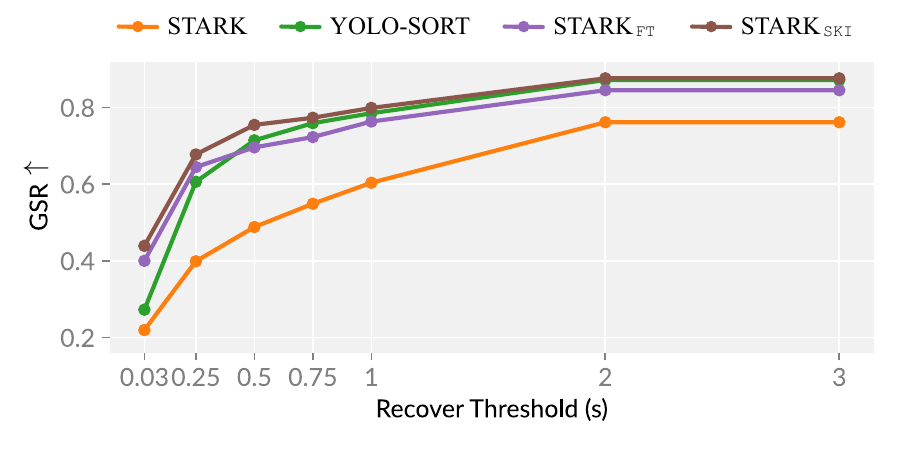}
  \caption{\textbf{Fraction of consistent skier tracking starting from the top.} This plot depicts the average fraction of consecutive frames in which the target skier is accurately localized before losing track, measured as the GSR score \cite{dunnhofer2023visual}. Various time thresholds in seconds are employed to assess the trackers' ability to recover from failures over time \cite{VOT2020}.} 
  \label{fig:gsr}
\end{figure}

\subsection{Tracking Impact}
\label{sec:hpe}
Skier localization plays a crucial role in high-level skiing performance analysis pipelines \cite{Stepec_2022_WACV,Ludwig_2022_WACV,wang2019ai,SkiPosePTZ}. To assess the impact of trackers in this context, we considered the 
2D body and equipment pose estimation task \cite{SkiPosePTZ,Ludwig_2023_WACV}. We focused on the AL and JP disciplines utilizing the Ski2DPose dataset \cite{SkiPosePTZ} and the YouTube Skijump dataset \cite{Ludwig_2023_WACV} respectively. These datasets provide sparse 2D key-point annotations for specific frames within a skiing video. 
We extracted tracking sequences referring to the same skier based on these annotations, and used a tracker to locate the skier in all the frames between the first and last occurrence of the annotations. The predicted boxes from the tracker were then used by a fine-tuned AlphaPose instance \cite{AlphaPose} to estimate 2D poses. We compared the predicted poses with the ground-truth annotations using the Percentage of Correct Key-points (\pck) and the Mean Per Joint Position Error (\mptje) \cite{SkiPosePTZ}. 
We also employed the metrics discussed in Section \ref{sec:trpm} to compare the tracker's predictions with the boxes extracted from the ground-truth key-point coordinates. This analysis allowed us to correlate the tracking and pose estimation accuracies.

\begin{table}[t]
\fontsize{7}{8}\selectfont
	\centering
	\caption{\textbf{Detector-based initialization.} The \fscore\ of different trackers is compared in terms of a ground-truth-based (left of $\rightarrow$) and detection-based initialization (right of $\rightarrow$).
	}
	\label{tab:detinit}
	\setlength\tabcolsep{.08cm}
	\begin{tabular}{l | c  c  c  c }
		\toprule
		  Tracker & AL & JP & FS & All  \\
		
            \midrule

            STARK & 0.552 $\rightarrow$ 0.544 &  0.603 $\rightarrow$ 0.590 & 0.596 $\rightarrow$ 0.497 & 0.584 $\rightarrow$ 0.544 \\
            \yolotr & 0.798 $\rightarrow$ 0.798 & 0.818 $\rightarrow$ 0.818 &  0.603 $\rightarrow$ 0.588 & 0.740 $\rightarrow$ 0.735 \\
            \starkft & 0.853 $\rightarrow$ 0.850  & 0.880 $\rightarrow$ 0.879 &  0.721 $\rightarrow$ 0.698 & 0.818 $\rightarrow$ 0.809\\
            \algoname & 0.868 $\rightarrow$ 0.870  & 0.896 $\rightarrow$ 0.897 & 0.742 $\rightarrow$ 0.696  & 0.835  $\rightarrow$ 0.821  \\

		\bottomrule		
\end{tabular}
\end{table}

\begin{figure}[t]
\centering
  \includegraphics[width=\linewidth]{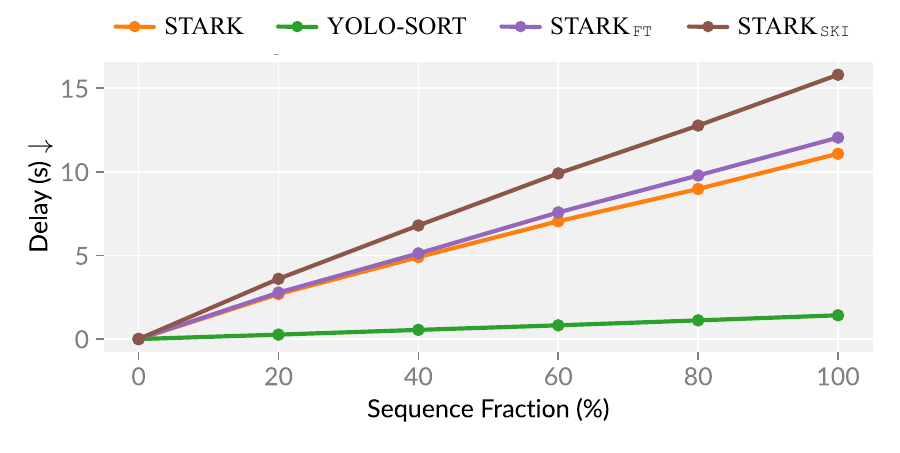}
  \caption{\textbf{Waiting time to obtain skier localizations.} The plot illustrates at various fractions of an MC sequence the average time that has to be waited to get the bounding-boxes from the trackers. \yolotr\ demonstrates the highest efficiency, with minimal delay compared to the actual happening of the skiing performance.} 
  \label{fig:eff}
\end{figure}

\section{Results}

\paragraph{General Remarks.}
Table \ref{tab:perdisciplinelt} presents the tracking performance of the selected trackers on the date-based test set of \datasetname. Among generic object trackers, STARK results the best. Its performance is comparable to the one achieved on traditional long-term benchmark datasets \cite{VOT2021,LaSOT}. These results indicate that generic object trackers struggle to generalize to the application settings represented by \datasetname. 
on the other hand, skier-specific trackers perform significantly better, with \starkft\ improving STARK's \fscore\ by 40\%. In this score, \algoname\ achieves an additional 2\% increase over \starkft. Comparing these findings with the results obtained for the SC setting (available in Table \ref{tab:perdisciplinest} of Appendix \ref{sec:addres}), we observe that camera shot-cuts introduce challenges that adversely affect tracking performance. Figure \ref{fig:qualex} visually illustrates the tracking accuracy of the top methods STARK, \yolotr, \starkft, \algoname. Overall, we can state that skier-specific trackers show promising performance for the application in real-world, especially in videos acquired by the same camera.

\begin{figure}[t]
\centering
  \includegraphics[width=\linewidth]{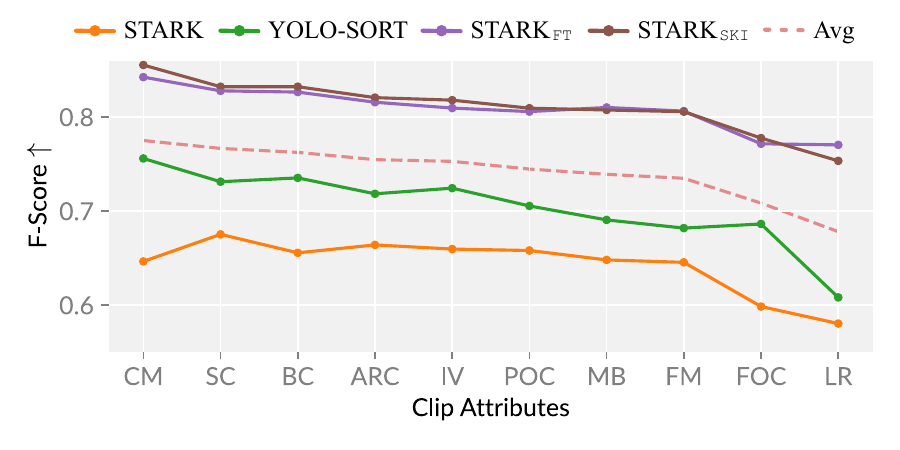}
  \caption{\textbf{Tracking performance based on visual attributes.} This plot reports the \fscore\ for the different attributes used to characterize the single-camera (SC) clips. We observe that the low resolution (LR), the full occlusion (FOC), and the fast motion (FM) of skiers are the most difficult situations to address.} 
  \label{fig:attributes}
\end{figure}

\begin{figure}[t]
\centering
  \includegraphics[width=\linewidth]{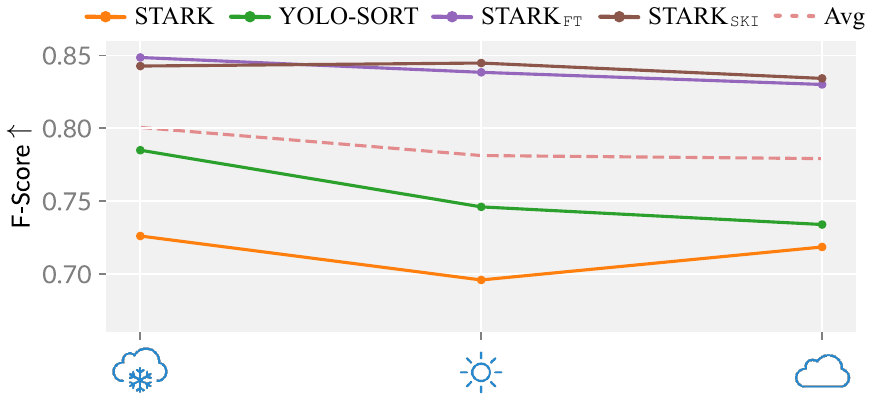}
  \caption{\textbf{Impact of the weather conditions on tracking.} This plot displays the \fscore\ based on the weather conditions (harsh, sunny, cloudy) characterizing the SC clips. In general, the tracking accuracy is not influenced by different weather conditions.} 
  \label{fig:weatherplot}
\end{figure}

\begin{table}[t]
\fontsize{7}{8}\selectfont
	\centering
	\caption{\textbf{Tracking after different training conditions.} The table reports \starkft's \fscore\ under different generalization conditions. Generalizing to unseen athletes is more challenging compared to unseen courses or newer skiing performances.}
	\label{tab:splitseval}
	\setlength\tabcolsep{.32cm}
	\begin{tabular}{l | c  c  c  c }
		\toprule
		  Generalization Condition & \multirow{1}{*}{AL} & \multirow{1}{*}{JP} &  \multirow{1}{*}{FS} & \multirow{1}{*}{All}  \\
		
            \midrule

            New performances & 0.853  & 0.880 &  0.721 & 0.818 \\
            Unseen athletes & 0.854 & 0.890 & 0.682 & 0.809 \\
            Unseen courses & 0.861 & 0.917 & 0.808 & 0.862 \\

		\bottomrule		
\end{tabular}
\end{table}

\begin{table*}[t]
\fontsize{8}{9}\selectfont
	\centering
	\caption{\textbf{Impact of trackers on high-level skiing performance understanding tasks.} We report the impact of the top trackers' predictions in the task of 2D  body and equipment pose estimation for the alpine skiing (AL) and ski jumping (JP) disciplines. As can be expected, more accurate trackers lead to a more accurate pose prediction in general. (GT boxes were extracted from the annotated pose key-points).
	}
	\label{tab:applications}
	\setlength\tabcolsep{.14cm}
	\begin{tabular}{c | c  | l  | c | c c c c | c }
		\toprule
		  Discipline & Dataset & Task & Metric & STARK & \yolotr & \starkft & \algoname & GT box \\
		
            \midrule

             \multirow{2}{*}{AL} & \multirow{2}{*}{Ski2DPose} & Tracking & \fscore & 0.751 & 0.830 & 0.848 & 0.849 & - \\
             &  & Pose Estimation & \pck\ / \mptje & 0.573 / 0.059 & 0.685 / 0.034 & 0.694 /  0.033 & 0.686 / 0.033 & 0.682 /  0.036 \\

            \midrule

            \multirow{2}{*}{JP} & \multirow{2}{*}{YouTube Skijump} & Tracking & \fscore & 0.670 & 0.748 & 0.768 & 0.775 & - \\
             &  & Pose Estimation & \pck\ / \mptje & 0.516 / 0.029 & 0.598 / 0.026 & 0.574 /  0.023 & 0.596 /  0.026 & 0.571 /  0.026 \\

		\bottomrule		
\end{tabular}
\end{table*}

\paragraph{In-depth Analysis.}
In this section, we present a deeper analysis of the application domain on the four most accurate methods presented in Table \ref{tab:perdisciplinelt}.

JP receives the best tracking performance among the skiing disciplines, followed by AL, while FS presents more challenges. This may be attributed to the complex poses athletes perform in such a discipline, and to the presence of other skiers in the videos as occurs in the  sub-disciplines of ski cross and dual moguls.

Figure \ref{fig:gsr} displays the proportion (as \gsr\ scores) of the skiing performance that the trackers are able to consistently cover before losing track of the target skier. Overall, \algoname\ demonstrates the most promising performance. With a recovery time of 1 second or longer, fractions scores exceed 80\%. However, shorter recovery time thresholds result in shorter coverages. With a threshold of 1 frame (i.e., 0.03s), \algoname\ achieves successful continuous tracking for around the first 40\% of the athlete's performance.

Table \ref{tab:detinit} presents the impact, as reflected in the change in \fscore, of using the YOLOX detector \cite{YOLOX,YOLOv5} to initialize the tracker. It is observed that the generic STARK is more sensitive to initialization noise. Conversely, skier-specific methods exhibit greater robustness to such initialization, with \algoname\ maintaining good scores in general. Notably, the detection initialization has a more significant effect on the FS discipline, likely due to the initialization to a wrong skier in multi-athlete sub-disciplines.

In terms of running speed, Figure \ref{fig:eff} analyzes the time taken by trackers to provide skier localization at different fractions of the observed skiing performance. \yolotr\ gives the best efficiency, offering minimal delay compared to the unfolding of the skiing performance. \algoname\ is the least efficient tracker, it accumulates delay while processing all the video frames and finally provides the last localization of the skier over 15 seconds after his/her performance has ended.

By analyzing the \fscore\ per the visual attributes characterizing the SC clips, as reported by Figure \ref{fig:attributes}, we notice that the small size (LR), the full occlusion (FOC), and the fast motion (FM) of skiers are the conditions that determine a performance drop. On the other hand, the camera motion (CM), the scale change (SC), and the background clutter (BC) consist in situations better addressed by the trackers.

Figure \ref{fig:weatherplot} shows that trackers work generally at the same accuracy level under different weather conditions. These results demonstrate that such methodologies can be used reliably under  challenging image conditions.

Table \ref{tab:splitseval} presents the performance of \starkft\ trained on different splits representing various real-world usage conditions. The results indicate that generalizing to unseen athletes is the most challenging application case. Generalizing to skiing performances occurring after the training ones also proves to be a demanding application condition. Unseen courses instead pose fewer difficulties for generalization.

Finally, Table \ref{tab:applications} presents the impact of the trackers on the 2D skier pose estimation tasks described in Section \ref{sec:hpe}. Generally, we observe that employing skier-specific trackers improves the skier tracking results as well as the pose estimation results. For AL on Ski2DPose, \algoname\ and \starkft\ have nearly the same impact on AlphaPose, while \algoname\ and \yolotr\ have comparable impact in JP as represented by the YouTube Skijump dataset. Across disciplines, \algoname\ results the best generalizing tracker by tracking and impact scores. Overall, these results show that the trackers' performances on \datasetname\ reflect the impact on high-level skiing understanding tasks.
It is worth mentioning that these results are obtained with the limited annotations present in the respective small-scale datasets. We hypothesize the relation with the \datasetname's results to become more evident on more densely-labeled datasets.

\section{Conclusions}
This paper presented a comprehensive study on tracking skiing athletes in monocular multi-camera broadcasting videos. Through the evaluation of established and newly introduced methodologies on the released dataset \datasetname, the study revealed that fine-tuned application-specific deep learning-based algorithms demonstrate consistent tracking performance and promising applicability throughout a skier's performance. These trackers exhibit robustness under various conditions such as challenging weather, fast camera motion, and background clutter, and they generalize well to new locations of application.
However, the study also identified certain limitations the prevent the methods to be perfect. Challenges arise in maintaining a continuous per-frame reference to the target skier across camera shot-cuts, in accurately localizing the skier in the presence of distractors, small appearance, occlusion, and fast motion. Additionally, the generalization to unseen athletes poses particular difficulties. Top-performance trackers should be also improved in their efficiency.
Future work will focus on addressing these limitations. Solutions may involve refining skier-specific tracking methods, improving generalization, and developing strategies to better integrate with high-level skiing performance understanding modules.

\paragraph{\footnotesize Acknowledgments.} 
{\footnotesize
Research supported by the project between the University of Udine and the organizing committee of EYOF 2023 Friuli-Venezia Giulia.}

{\small
\bibliographystyle{ieee_fullname}
\bibliography{egbib}
}

\clearpage

\twocolumn[{%
\centering
\vspace{1em}
{\Large \textbf{Tracking Skiers from the Top to the Bottom}} \\
\vspace{.5em}
{\large  Supplementary Document} \\
\vspace{1em}
Matteo Dunnhofer, Luca Sordi, Niki Martinel, Christian Micheloni \\
Corresponding author e-mail: \href{mailto:matteo.dunnhofer@uniud.it}{\texttt{matteo.dunnhofer@uniud.it}}
\vspace{2.5em}
}]

\appendix

\section{Further Details on \datasetname}
\label{sec:datadet}

This section provides additional information about the data contained in \datasetname\ as well as further motivations behind its construction.

To avoid confusion, we state that, in the scope of this paper, a skiing course is considered as a path or track down a mountain slope that an athlete should follow to complete his/her performance. It should not be confused with a course taken to learn how to ski.

\subsection{Bounding-box Representation}
As stated in the main paper, the motivation behind the employment of bounding-boxes is grounded on the fact that such a representation is sufficiently informative for the computational processes performed by higher-level skiing performance understanding tasks \cite{SkiPosePTZ,Ludwig_2022_WACV,Ludwig_2023_WACV,Stepec_2022_WACV}. The aforementioned pipelines simply require a rectangle highlighting the area covered by the skier's appearance. Compared to the more complex segmentation masks \cite{VOT2020, VOT2021}, the four-value representation of bounding-boxes demands less computational resources, thus enabling the development of more efficient methods.
Additionally, the choice of including the appearance of the skiing equipment within the labeled bounding-box is guided by the common working mechanism of the aforementioned solutions, which necessitate a bounding-box encompassing both the athlete's body and equipment.

\begin{table*}[t]
\fontsize{7}{8}\selectfont
	\centering
	\caption{\textbf{Selected sequence attributes associated to single-camera (SC) clips.} This table gives the formal definition of the selected clip attributes according to previous research in generic visual object tracking \cite{OTB,LaSOT,dunnhofer2023visual}. On a side, we give an interpretation of each definition w.r.t. our application domain.}
	\label{tab:attrdesc}
	\begin{tabular}{m{3em} | m{31em} | m{31em} }
		\toprule
		Attribute & Definition & Application-specific Interpretation \\
		\midrule
        CM & \underline{Camera Motion}: an abrupt camera motion can be seen in the video clip. & The camera operator moves the camera fast to keep the skier in the field of view. \\
		SC & \underline{Scale Change}: the ratio of the bounding-box area of the first and the current frame is outside the range [0.5, 2]. & The size of a skier's appearance changes considerably during the video (e.g. by zooming in/out on the target). \\
        BC & \underline{Background Clutter}: the target has a similar appearance w.r.t. the surrounding background. & The appearance of the athlete's suit and equipment confounds with the elements in the background. \\
		ARC & \underline{Aspect Ratio Change}: the ratio of the bounding-box aspect ratio of the first and the current frame is outside the range [0.5, 2]. & The ratio between the height and width of the athlete changes (e.g. due to complex body poses). \\
		IV & \underline{Illumination Variation}: the area of the target bounding-box is subject to light variation. & The appearance of the target skier changes due to particular lightning conditions (e.g. passing through slope areas under shadow). \\
        POC & \underline{Partial Occlusion}: the target is partially occluded in the video. & Part of the skier is hidden by another item (e.g. by a gate in AL). \\
        MB & \underline{Motion Blur}: the target region is blurred due to target or camera motion. & The appearance of the skier is blurred due to its fast motion or the fast motion of the camera. \\
		FM & \underline{Fast Motion}: the target bounding-box has a motion change larger than its size. & The skier moves fast during the descent on the course. \\
		FOC & \underline{Full Occlusion}: the target is fully occluded in the video. & The skier is completely occluded by another item in the field of view (e.g. by a kicker in FS). \\
		LR & \underline{Low Resolution}: the area of the target bounding-box is less than 1000 pixels in at least one frame. & The skier appears small due to a low level of camera zoom. \\
		
		\bottomrule		
\end{tabular}
\end{table*}
\subsection{Details on the Visual Attribute Labels}
\begin{figure}[t]
\centering
  \includegraphics[width=\columnwidth]{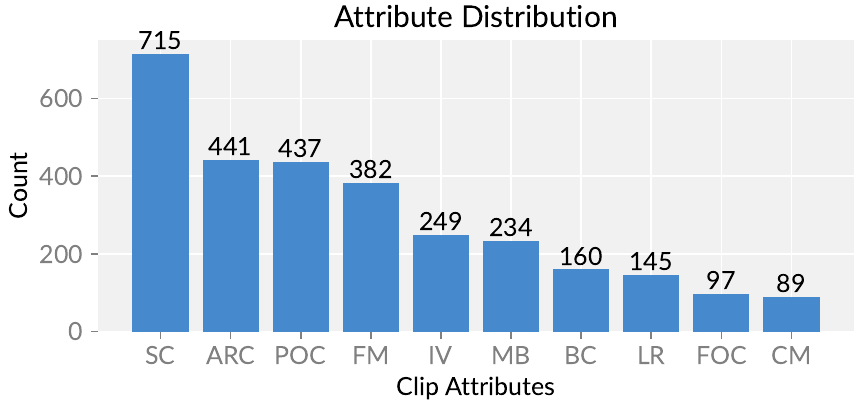}
  \caption{\textbf{Distribution of the clip attributes.} The plot shows the number of single-camera (SC) clips associated with each of the attributes introduced to characterize the visual variability of the target, as in \cite{OTB,LaSOT,UAV123,dunnhofer2023visual}. The application domain of skiing videos presents a large number of scale changes (SC), followed by a substantial number of partial occlusions (POC), changes in the aspect ratio (ARC), and fast motions (FM).
  } 
  \label{fig:attrdist}
\end{figure}

Table \ref{tab:attrdesc} presents the description of the attributes assigned to the SC clips. The attributes have been introduced to cluster the tracking performance depending on the visual variability events occurring on the target object. This evaluation approach of assigning per-video labels is well-established in the visual object tracking community \cite{OTB,NfS,dunnhofer2023visual,UAV123,LaSOT,GOT10k} and was shown to be sufficiently robust to estimate the trackers' performance in particular scenarios. 
Among the many attributes present in the literature, we selected 10 that well represent the variability of the skiing domain. The labels have been associated with SC clips of the date-based training-test split because the SC experimentation setting allows a tracker to cover the situations happening during the skier's descent in a more complete and consistent way \cite{VOT2020,dunnhofer2023visual}. 
Figure \ref{fig:attrdist} shows the distribution of the SC clips according to the labels.
In \datasetname, the labels SC, ARC, FM, and LR, have been assigned by an automatic procedure as described by \cite{OTB,LaSOT}. The presence of situations identified by the other attributes has been visually assessed and annotated by our research team. 

\begin{figure}[t]
\centering
  \includegraphics[width=\linewidth]{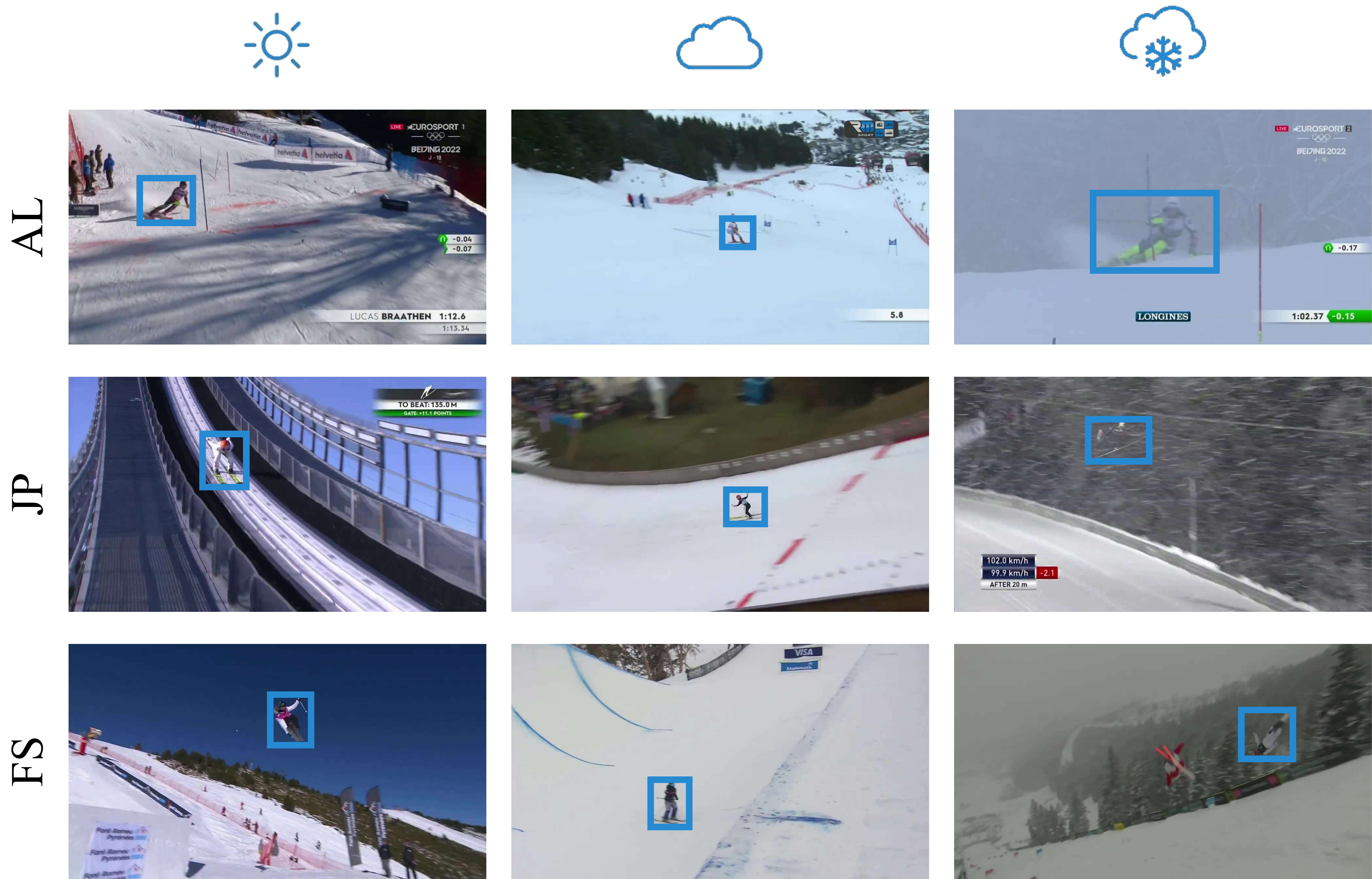}
  \caption{\textbf{Winter weather conditions.} Skiing takes place in winter environments, subjecting athletes to extreme weather conditions that introduce unique image characteristics when captured on camera. For instance, sunny conditions can create shadows, resulting in significant variations in target illumination. Cloudy weather leads to "flat light" conditions, reducing image contrast, while snowfall or rain further diminishes visibility. The \datasetname\ includes weather condition labels for each MC video.} 
  \label{fig:weathercond}
\end{figure}

\subsection{Details on the Weather Labels}
The weather labels have been associated with each MC video because the weather condition generally remains the same across all the location in which the skiing competition takes place. The labeling of the conditions was performed by our team by analyzing the condition visible in the video. Such a label was also checked to match the one reported on the official result list available on the FIS database \cite{FISSKI}. The labeling generated the following weather labels: ``Clouds'', ``Fog'', ``LowClouds'', ``MostlyCloudy'', ``Overcast'', ``PartlyCloud'', ``Raining'', ``Snowing'', ``Sunny'', ``Clear''. In order to have a larger number of samples for the experiments, such labels have been clustered into three categories: ``Sunny'', ``Cloudy'', and ``Harsh'' weather. Figure \ref{fig:weathercond} gives examples of the image condition such weather condition cause. In total, \datasetname\ provides 191 videos associated with ``Sunny'', 66 with ``Cloudy'', and 43 associated with ``Harsh''. After the date-based training-test split, the test set used to compute the results in Figure \ref{fig:weatherplot} has 80 ``Sunny'', 26 ``Cloudy'', and 14 ``Harsh'' videos.

\begin{table*}[t]

\fontsize{8}{9}\selectfont
	\centering
	\caption{\textbf{Statistics of \datasetname's training and test splits.} The following table reports some statistics of the three split that have been created to evaluate the capability of learning-based trackers in generalizing to different application conditions. For generalizing to new performances, the date associated to the videos has been used as splitting condition; for the generalization to unseen athletes, the athlete IDs; to generalize to unseen courses, the course's location information. 
 }
	\label{tab:splitsstats}
	\setlength\tabcolsep{.4cm}
	\begin{tabular}{l | c  c  | c  c | c c }
		\toprule
		Generalization Condition  & \multicolumn{2}{c|}{New performances} & \multicolumn{2}{c|}{Unseen athletes} & \multicolumn{2}{c}{Unseen courses} \\
        Split & Train & Test & Train & Test & Train & Test \\
		\midrule
		\# MC videos & 180 & 120 & 176 & 124 & 182 & 118 \\
            \# SC videos & 1215 & 804 & 1151 & 868 & 1238 & 781 \\
		\# frames & 212793  & 140185 & 202052 & 150926 & 213093 & 139885 \\
        avg MC video seconds & 39 & 39 & 38 & 41 & 39 & 40 \\
        avg SC video seconds & 5.8 & 5.8 & 5.9 & 5.8 & 5.7 & 6.0 \\
            \# sub-disciplines & 11 & 11 & 11 & 11 & 11 & 11 \\
            \# athletes & 136 & 90 & 118 & 78 & 138 & 95 \\
            \# athlete genders (M, W) & (84, 52) & (53, 37) & (70, 48) & (47, 31) & (83, 55) & (56, 39) \\
            \# athlete nationalities & 21 & 22 & 22 & 21 & 23 & 20 \\
            \# locations & 124 & 61 & 122 & 92 & 99 & 62 \\
            \# location countries & 21 & 20 & 23 & 23 & 22 & 20 \\
		
		\bottomrule		
\end{tabular}

\end{table*}

\paragraph{Details on the Training-Test Splits.}
Table \ref{tab:splitsstats} shows some statistics of the videos present in the three different training-test splits generated to train and test skier-specific trackers under different application conditions. The splits have been generated to maintain a balanced distribution across the skiing disciplines and sub-disciplines while aiming to keep condition-specific disjoint partitions and respect as close as possible a 60-40 ratio.

\section{Details about the Trackers}
\label{sec:skiertrackers}
In this section, we give some more details of the implementation of the selected trackers. 

\subsection{Generic Object Trackers} 
The generic object trackers have been selected to be representative of the state-of-the-art solutions in the years between 2010-2023. They have been implemented by exploiting the code originally provided by the authors along with pre-trained weights. The original hyper-parameter values leading to the best and most likely generalizable instances of all the trackers have been set. Those trackers that do not output a confidence score, were modified to return an always-confident score of 1.0.

\subsection{Skier-specific Trackers} 

\paragraph{\yolotr.}
The \yolotr\ tracker implements a tracking-by-detection approach inspired by multiple object tracking \cite{SORT,dendorfer2021motchallenge}. 
At each frame of a video, this baseline first detects skiers with an YOLOX instance \cite{YOLOX}\footnote{\url{https://github.com/ultralytics/yolov5}} and then exploits the Simple Online and Realtime Tracking method (SORT) \cite{SORT} to associate the new detections with previously memorized tracklets. The YOLOX instance was trained on all the frames and the associated bounding-box annotations of \datasetname's training set defined by the date-based split, by mostly default hyper-parameters. The only changes made are relative to the batch size, set to 16, and the number of training epochs, set to 25. 10\% of the training videos were considered to build the set of validation images. The model instance achieving the highest Average Precision (AP) on such a subset was retained for inference during tracking. The SORT module is initialized in the first frame with the given skier's bounding-box. At every other time step, it is given in input all the detections given by YOLOX and returns a new set of tracks. As output, we retain the bounding-box associated to the track initialized in the first frame.

\paragraph{\starkft.}
The \starkft\ baseline implements a fine-tuned version of the generic object tracker STARK (STARK-ST50) \cite{Stark}. To implement this tracker we exploited the publicly available code\footnote{\url{https://github.com/researchmm/Stark}} and adapted the model's tracking ability by fine-tuning on \datasetname's training set, according to STARK's original training strategy. Mostly default hyper-parameters have been kept, except for the number of epochs in stage-one training, which has been set to 200.

\begin{algorithm}[t!]
    \begin{algorithmic}[1]
    \caption{Pseudo-code of the procedure implemented by the proposed \algoname\ while running on a video.}
    \label{algo:algoname}
    \fontsize{8}{9}\selectfont
    \State \emph{// Consider video $\video$ and ground-truth box $\bbox_0$}
    \State \emph{// Trackers initialization}
    \State Initialize \starksc with $\frame_0$ and $\bbox_0$
    \State Initialize \starkft with $\frame_0$ and $\bbox_0$
    \State $t \gets 1$
    \Repeat
    
    \State $\bbox_t, \conf_t \gets $ Run \starksc\ on $\frame_t$ 

    \If{$\conf_t \leq 0.5$} 
        \State$\bbox_t, \conf_t \gets $ Run \starkft\ on $\frame_t$
        \If{$\conf_t > 0.5$} \State \emph{// \starksc\ re-initialization}
            \State Re-initialize \starksc\ with $\frame_t$ and $\bbox_t$
        \EndIf
    \Else
        \State \emph{// Compute bounding-box for \starkft\ relocalisation}
        \State $S \gets \frac{H}{5.0}$ // 5.0 is \starkft\ search area's factor
        \State $x_t^{*} \gets clip(x_t, \frac{H}{2}, W - \frac{H}{2})$ 
        \State $y_t^{*} \gets \frac{H}{2} - \frac{S}{2}$ 
        \State $\bbox_t^{(R)} \gets [x_t^{*}, y_t^{*}, S, S]$
        
        \State Use $\bbox_t^{(R)}$ to reset \starkft's box used to compute the search area location
    \EndIf

    \State Return $\bbox_t, \conf_t$ as output for $\frame_t$
    \State $t \gets t + 1$
    \Until $t = T$
    \end{algorithmic}
\end{algorithm}
\paragraph{\algoname.}
The pseudo-code of the procedure implemented by the proposed skier-optimized tracking baseline \algoname\ is given in Algorithm \ref{algo:algoname}. The procedure is composed of two skier-specific instances of STARK (STARK-ST50) \cite{Stark}. The first one, which we refer to as \starksc, is a modified version of STARK that, at every frame, computes the target bounding-box by exploiting a higher-resolution search area located around the previous target location. This is achieved by reducing the search area factor from the original value of 5.0 to 3.0 (we determined the value 3.0 by experiments) and fine-tuning as done for \starkft. In this way, we reduce the amount of background information present in the search area, thus increasing the resolution of the target skier's appearance and making the tracker predict more accurate bounding-boxes during single-camera tracking. Given the more limited search area, \starksc\ performs better just in such conditions where the target and camera motion are stable and consistent across consecutive frames. In the other cases, i.e. in those frames where \starksc\ is not confident in tracking the target (lines 8-13 of Algorithm \ref{algo:algoname}), we exploit a \starkft\ instance configured as described in the previous paragraph. This instance keeps the original search factor with a value of 5.0 and thus is able to look for the target in a larger frame area. The execution of this \starkft's instance is generally triggered after a camera shot-cut and during the complete occlusion of the target. We empirically found beneficial to set the search area size of this instance to match the frame's height, by modifying the bounding-box values that are used to compute the search area at the next frame (lines 16-20 of Algorithm \ref{algo:algoname}). The position of such a box is set to be the latest confident box position predicted by \starksc, clipped to make the search area not fall outside of the frame. Whenever \starkft\ finds confidently the target again, its predicted bounding-box and the respective frame are used to re-initialize \starksc. We found the re-initialization to work better than just relocating \starksc\ on the \starkft's predicted bounding-box.

\section{Details on the Evaluation}
\label{sec:expdet}
In this section, we explain and motivate in more detail the evaluation procedures implemented.

\paragraph{Evaluation Protocols.}
As mentioned in the main paper, to run a tracker, we employed the OPE protocol introduced in \cite{OTB} which implements the most realistic way to run a tracker in practice.
The protocol consists of two main stages: (i) initializing a tracker with a bounding-box of the target in the first frame of the video; (ii) letting the tracker run on
 every subsequent frame until the end and recording predictions to be considered for the evaluation. To obtain performance scores for each sequence, predictions and ground-truth bounding-boxes are compared according to some distance measure. The overall scores
 are obtained by averaging the scores achieved for every sequence.
 As in the default OPE, we use  the ground-truth bounding-box for initialization to evaluate the trackers in the best possible conditions, i.e. when accurate information about the target is given. However, many deployment conditions do not allow human labeling but instead require a completely automatic athlete localization system (e.g. real-time skiing performance analysis during broadcasting). To evaluate trackers in similar conditions, we use an object detector to predict the initial skier bounding box. 
 Thus, we consider a version of the OPE protocol where each tracker is initialized in the first frame in which the YOLOX detector's \cite{YOLOX,YOLOv5}, fine-tuned for skier localization, provides a bounding-box prediction with confidence score $\geq$ 0.5. 
 The fine-tuning of this detector was performed in the same way as for \yolotr\ mentioned before. 

 \paragraph{Performance Measures.}
To quantify the distance between the predicted and temporally-aligned ground-truth bounding-boxes, we used different measures. 
As general tracking accuracy indicators, we employed the metrics defined by \cite{Lukezic2020} for long-term tracking problems: Precision, Recall, and F-Score. 
Due to the generally long video observation and presence of multiple occlusions, our problem of interest is related to such a research framework.
Now we explain the meaning of such metrics in relation to our application case. 
The Precision (\prec) measures the average amount of correctly tracked ground-truth bounding-boxes where the tracker is confident, with different thresholds used to determine the conditions of correct and confident prediction. In our case, the \prec\ score determines the average coverage of the skier's position on the portion of skiing performance observation on which the tracker is confident. For example, a \prec\ score of 0.8 tells that an algorithm correctly localizes the athlete for the 80\% of the bounding-box predictions that are given with high confidence.
The Recall (\recall) instead measures the average amount of correctly tracked ground-truth bounding-boxes, regardless of the tracker's confidence. In our context, such a score determines the average coverage of the position of the skier throughout the whole skiing performance. For instance, a \recall\ score of 0.8 gives that the algorithm correctly localizes the athlete for 80\% of the skiing performance appearing in the video.
The F-Score (\fscore) provides a single aggregating score that incorporates both the previous measures. The best value across the different confidence thresholds is retained.

In addition to those metrics, we exploited the Generalized Success Robustness (\gsr) \cite{dunnhofer2021first,dunnhofer2023visual} which reports the fraction of continuous successful tracking before the tracker is lost, measured as the temporal index of the first wrong prediction normalized by the number of frames in the video. In the context of this application domain, such a metric reports the percentage of continuous coverage of the skier's performance before the target is lost by the tracking algorithm. The original metric \cite{dunnhofer2023visual} is strict because it considers just the first wrong prediction to determine the tracker's failure time step. Other work \cite{VOT2020} suggested a softer version of such a measure. If the algorithm gets back to the target within a range of 10 consecutive frames, the tracking is resumed. Inspired by such a work, we evaluate the \gsr\ with several different temporal ranges to detect a failure, specifically 1 frame (0.03s), 7 frames (0.25s), 15 frames (0.5s), 22 frames (0.75s), 30 frames (1s), 60 frames (2s), and 90 frames (3s).

Finally, we assessed the computational efficiency of the trackers. This has been done by quantifying the time difference (in seconds) between the time stamp associated with each frame and the time instant on which the localization for the respective frame is given by an algorithm. Considering that sports performance analysis requires the processing of all the frames for a smooth and continuous understanding, a tracker that is slow will accumulate time while processing all the frames and delay its predictions. Thus, it becomes interesting to know how much time should be waited in order to obtain the localization, and how such delay grows during the online processing of the video. We give such a measurement in seconds with \timediff. 

\begin{table*}[t]
\fontsize{7}{8}\selectfont
	\centering
	\caption{\textbf{Overall and per-discipline results in the single-camera (SC) setting.} The \fscore, \prec, and \recall\ scores are presented for each studied algorithm. This setting is easier to tackle by all the algorithms in general. The different skiing discipline pose challenges to the trackers in the same way as in the multi-camera (MC) setting.
	}
	\label{tab:perdisciplinest}
	\setlength\tabcolsep{.24cm}
	\begin{tabular}{c | c   c  c  c c c c c c c c c | c c c}
		\toprule
		\rotatebox[origin=c]{90}{Discipline} &  \rotatebox[origin=]{90}{KCF} & 
  \rotatebox[origin=c]{90}{MOSSE} &\rotatebox[origin=c]{90}{FEAR} & 
  \rotatebox[origin=c]{90}{SiamRPN++} & \rotatebox[origin=c]{90}{GlobalTrack} &  
  \rotatebox[origin=c]{90}{LTMU} & \rotatebox[origin=c]{90}{SeqTrack} & \rotatebox[origin=c]{90}{KeepTrack} & \rotatebox[origin=c]{90}{OSTrack} &
        \rotatebox[origin=c]{90}{CoCoLoT} & \rotatebox[origin=c]{90}{MixFormer} & 
  \rotatebox[origin=c]{90}{STARK} & \rotatebox[origin=c]{90}{\yolotr} & \rotatebox[origin=c]{90}{\starkft} & \rotatebox[origin=c]{90}{\algoname} \\
		\midrule

        \multirow{3}{*}{All} & 0.294 & 0.367  &  0.564 & 0.583 & 0.592 & 0.642 & 0.645 & 0.654 &  0.663 & 0.681 & 0.686 & 0.703 &  \third{0.751} & \second{0.836} & \first{0.841} \\
         &  \smallresult{0.291} & \smallresult{0.363} & \smallresult{0.565} & \smallresult{0.583} & \smallresult{0.591} &  
         \smallresult{0.637} &
         \smallresult{0.639} & \smallresult{0.652} &  \smallresult{0.654} & \smallresult{0.676} & \smallresult{0.676} & \smallresult{0.698} & \smallresult{0.743} & \smallresult{0.827} & \smallresult{0.833} \\
         &  \smallresult{0.299} & \smallresult{0.376} & \smallresult{0.572} & \smallresult{0.592} & \smallresult{0.601} & \smallresult{0.651} &  \smallresult{0.681} & \smallresult{0.664} &  \smallresult{0.704} & \smallresult{0.694} & \smallresult{0.658} & \smallresult{0.717} & \smallresult{0.763} & \smallresult{0.854} & \smallresult{0.858} \\

        \midrule

        \multirow{3}{*}{AL} &  0.220 & 0.267 &   0.518 & 0.536 & 0.585 & 0.623 &  0.578 & 0.640 &  0.594 & 0.652 & 0.637 & 0.671 &  0.819 & 0.875 & 0.882 \\
         &  \smallresult{0.218} & \smallresult{0.265} & \smallresult{0.524} & \smallresult{0.542} & 
         \smallresult{0.595} &  
         \smallresult{0.622} & \smallresult{0.576} & \smallresult{0.641} &  \smallresult{0.590} & \smallresult{0.650} & \smallresult{0.634} & \smallresult{0.672} & \smallresult{0.814} & \smallresult{0.876} & \smallresult{0.886} \\
         & \smallresult{0.222} & \smallresult{0.269} &  \smallresult{0.516} & \smallresult{0.534} & \smallresult{0.578} & \smallresult{0.625} &  \smallresult{0.580} & \smallresult{0.640} & \smallresult{0.599} & \smallresult{0.655} & \smallresult{0.643} & \smallresult{0.672} & \smallresult{0.825} & \smallresult{0.876} & \smallresult{0.881} \\

        \midrule
        
        \multirow{3}{*}{JP}  & 0.389 & 0.487 &  0.641 & 0.663 & 0.677 & 0.702 &  0.747 & 0.705 &  0.763 & 0.738 & 0.765 & 0.761 & 0.855 & 0.899 & 0.907 \\
         &  \smallresult{0.388} & \smallresult{0.485} & \smallresult{0.645} & \smallresult{0.666} & \smallresult{0.677} & \smallresult{0.703} &  \smallresult{0.748} & \smallresult{0.707} &  \smallresult{0.760} & \smallresult{0.743} & \smallresult{0.762} & \smallresult{0.766} & \smallresult{0.850} & \smallresult{0.894} & \smallresult{0.901} \\
         &  \smallresult{0.390} & \smallresult{0.489} &  \smallresult{0.640} & \smallresult{0.660} & \smallresult{0.678} &  \smallresult{0.701} & \smallresult{0.747} & \smallresult{0.703} &   \smallresult{0.767} & \smallresult{0.734} & \smallresult{0.770} & \smallresult{0.758} & \smallresult{0.863} & \smallresult{0.907} & \smallresult{0.914} \\

        \midrule
        
        \multirow{3}{*}{FS}  & 0.274 & 0.347 &  0.531 & 0.550 & 0.514 &  0.599 & 0.612 & 0.617 &  0.633 & 0.654 & 0.654  & 0.676 & 0.578 & 0.734 & 0.735 \\
         & \smallresult{0.268} & \smallresult{0.338} &  \smallresult{0.525} & \smallresult{0.540} & \smallresult{0.502} & \smallresult{0.586} &  \smallresult{0.595} & \smallresult{0.607} &  \smallresult{0.612} & \smallresult{0.636} & \smallresult{0.633} & \smallresult{0.656} & \smallresult{0.566} & \smallresult{0.711} & \smallresult{0.713} \\
         & \smallresult{0.285} & \smallresult{0.370} &   \smallresult{0.560} & \smallresult{0.582} & \smallresult{0.548} & \smallresult{0.627} & \smallresult{0.647} & \smallresult{0.647} &  \smallresult{0.676} &  \smallresult{0.692} & \smallresult{0.698} & \smallresult{0.721} & \smallresult{0.601} & \smallresult{0.780} & \smallresult{0.780} \\

		\bottomrule		
\end{tabular}
\end{table*}

\subsection{Tracking Impact}
As stated in the main paper, the output of tracking is of paramount importance for many high-level modules that produce fine-grained skiing performance analyses \cite{Stepec_2022_WACV,wang2019ai,Ludwig_2023_WACV,Ludwig_2022_WACV,Dunnhofer_2023_CVPR}. Thus we evaluated the trackers based on the impact they have on the accuracy of such solutions. We think that the development of effective tracking methodologies should be driven not only by tracking-specific results but also by the contribution the algorithms bring in improving the accuracy of the overall system.

As an exemplar high-level skiing performance understanding tasks to evaluate tracker's impact, we focused on the problem of 2D pose estimation of skier body and equipment \cite{SkiPosePTZ,Ludwig_2023_WACV}. This task serves to obtain information regarding the position and orientation of specific human joints during exercises, and such an output is additionally exploited by even more high-level performance understanding modules such as 3D pose estimation \cite{SkiPosePTZ,CanonPose}.
 To estimate the image-level coordinates of a set of key-points that localize different parts of a skier's body (e.g. head, shoulders, hips, feet, etc.) and of particular points of interest of the skier's equipment (e.g. ski tips or tails), the available solutions \cite{SkiPosePTZ,Ludwig_2023_WACV} first run an object detector \cite{FasterRCNN,YOLO} to compute bounding-boxes for the athlete present in the input RGB image, and then crop image patches from such boxes that are successively given as input to a state-of-the-art deep neural network architecture (e.g. AlphaPose \cite{AlphaPose}) that predicts the key-point coordinates. 
Such a pose estimation network is trained by fine-tuning on ground-truth poses by exploiting input image patches extracted with bounding-boxes defined by the coordinates of the annotated key-points. 

The aforementioned studies \cite{SkiPosePTZ,Ludwig_2023_WACV} propose datasets of videos (with dedicated training and test sets) whose frames are sparsely labeled with the poses of body and equipment. The authors evaluate the proposed pipelines on such benchmarks but treat each frame as an independent image, and so during testing the object detector is run on every image before the pose estimation network. Considering the presence of videos, we use such datasets as a base for the evaluation of trackers as athlete localizers executed before the pose estimation step. Thus, we determine the tracker's impact by evaluating the accuracy of the pose estimation model, where the input of the latter is influenced by the output of the former.
After having trained an AlphaPose instance \cite{AlphaPose} on the original training images \cite{SkiPosePTZ,Ludwig_2023_WACV}, we evaluate its accuracy on the test frames by inputting it with a patch extracted from a tracker's box prediction. The evaluation of the pose estimator is done through: the Percentage of Correct Keypoints (\pck) which measures the number of predicted key-points, normalized by the number of all key-points \cite{SkiPosePTZ}, having a pixel distance lower than the 50\% of the ground-truth-based head-neck distance; and the Mean Per Joint Position Error (\mptje) which measures the normalized pixel distance between predicted and corresponding ground-truth key-points \cite{SkiPosePTZ}. The tracker's bounding-boxes are obtained by implementing the OPE protocol on the sequence of frames in between the first and the last pose annotation occurrences that refer to the same athlete. Indeed, we obtain boxes's top-left and bottom-right vertices by considering the lowest and greatest values in the key-points coordinates. The first bounding-box is considered for tracker initialization, while the others are for prediction evaluation.
We respect the original training-test separations \cite{SkiPosePTZ,Ludwig_2023_WACV}. For testing alpine skiing (AL) pose estimation on the Ski2DPose dataset, we used 11 video clips related to the 150 pose annotated images, while for pose estimation in ski jumping (JP) on the YouTube Skijump dataset we used 19 videos referring to the 118 annotated test images.

For the implementation and training of the AlphaPose instance \cite{AlphaPose}, we employed the Alphapose v0.6 framework.\footnote{\url{https://github.com/MVIG-SJTU/AlphaPose}} 
Specifically, we conducted two separate fine-tuning based on the ResNet50 model for Ski2DPose and YouTube Skijump. Both training sessions run for 250 epochs, employing a batch size of 32 and a learning rate of 0.001 decreased by a 0.1 factor every 70 epochs. During both training and testing, in the computation of the input image crop, a padding of 20\% was added to the dimensions of the available bounding-box.

\subsection{Implementation Details}
All the code used for our study was implemented in Python and run on a machine with an Intel Xeon E5-2690 v4 @ 2.60GHz CPU, 320 GB of RAM, and 8 NVIDIA TITAN V GPUs.

\section{Additional Results}
\label{sec:addres}
This section reports the results of additional experiments we conducted.

Table \ref{tab:perdisciplinest} presents the performance achieved by the selected trackers in the case of skier tracking on SC videos. This setting is more similar to the problem of short-term visual object tracking \cite{OTB,Lukezic2020} where the duration of the videos is shorter and they are captured by the same video camera (no camera shot-cuts are present). Application-wise, the conditions of SC tracking align with: the broadcasting requirements of skiing performance replay where just a specific section of the skiing performance is captured and played again; training processes where a trainer captures a specific section of the ski track/course with a smartphone for later video analysis. From the table, we observe that tracking a skier without camera-shots results easier in general. Generic object trackers show a larger improvement by tracking on SC videos than on MC ones. However, their tracking accuracy still remains lower than the skier-specific methods. Regarding the skiing disciplines, we notice that FS videos still cause the major drop in the overall performance.

\subsection{Videos}

Videos showing qualitative results of the top performing algorithms on \datasetname\ are available at \url{https://www.youtube.com/watch?v=Aos5iKrYM5o}.

\end{document}